\pdfoutput=1
\documentclass[10pt,twocolumn,letterpaper]{article}

\usepackage[pagenumbers]{wacv}
\usepackage{amsmath}
\usepackage{amssymb}
\usepackage{graphicx}
\usepackage{booktabs}
\usepackage{array}
\usepackage{multirow}
\usepackage{subcaption}

\newcommand{\Rset}{\mathbb{R}}

\newcommand{\bv}{\mathbf{v}}
\newcommand{\bt}{\mathbf{t}}
\newcommand{\bz}{\mathbf{z}}
\newcommand{\bW}{\mathbf{W}}
\newcommand{\bb}{\mathbf{b}}
\newcommand{\bA}{\mathbf{A}}
\newcommand{\bB}{\mathbf{B}}
\DeclareMathOperator{\LN}{LN}
\DeclareMathOperator{\GELU}{GELU}

\definecolor{linkblue}{rgb}{0.21,0.49,0.74}
\usepackage[pagebackref,breaklinks,colorlinks,allcolors=linkblue]{hyperref}

\title{Preserving the Manifold: Frozen Expert Alignment\\ for Label-Efficient SAR Scene Understanding%
\thanks{A shorter version of this work was presented at the ECCV 2026 Workshop
\emph{TerraBytes II: Towards Global Datasets and Models for Earth Observation} (non-archival).}}

\author{Eli A. Underwood \qquad I. Taylor McKechnie \qquad Leslie M. Collins\\ Duke University}

\begin{document}
\maketitle

\begin{abstract}
Cross-modal alignment of synthetic aperture radar (SAR) imagery to natural language
is constrained by two distinct failures.
The first is modality mismatch: text encoders pre-trained on optical imagery,
such as CLIP~\cite{Radford2021CLIP}, cannot be efficiently fine-tuned on SAR,
because active microwave backscatter is physically incompatible with passive optical
reflectance and the available SAR caption datasets are orders of magnitude too small
to bridge this gap without substantially degrading the learned semantic manifold.
The second is physical forgetting: SAR encoders pre-trained on large-scale
self-supervised objectives encode radar physics that are progressively erased
when the encoder is fine-tuned for text alignment on limited caption data.
A single design principle addresses both failures: keep both encoders frozen and learn only
a constrained, structure-preserving bridge.
Our Frozen Expert framework aligns a frozen CROMA~\cite{NEURIPS2023_11822e84}
SAR encoder with a frozen CLIP text encoder via a lightweight constrained residual bridge.
We formally define physical forgetting via Procrustes disparity and show our bridge
preserves SAR geometry significantly better than fine-tuning (disparity: 0.210
vs.\ 0.351).
We validate generality across multiple SAR backbones and text encoders.
On zero-shot SAR scene classification, our approach outperforms fine-tuned baselines
using up to 8$\times$ less training data:
DFC2020 (+10.3\% R@1, +15.5\% mAP),
EuroSAT-SAR (+15.7\% R@1, +12.2\% mAP),
BigEarthNet-MM (+14.7\% R@1).
These results establish frozen manifold alignment as a principled, data-efficient
alternative to fine-tuning in high-disparity cross-modal settings.
\end{abstract}

\section{Introduction}
\label{sec:intro}

\begin{figure}[t]
  \centering
  \includegraphics[width=\linewidth]{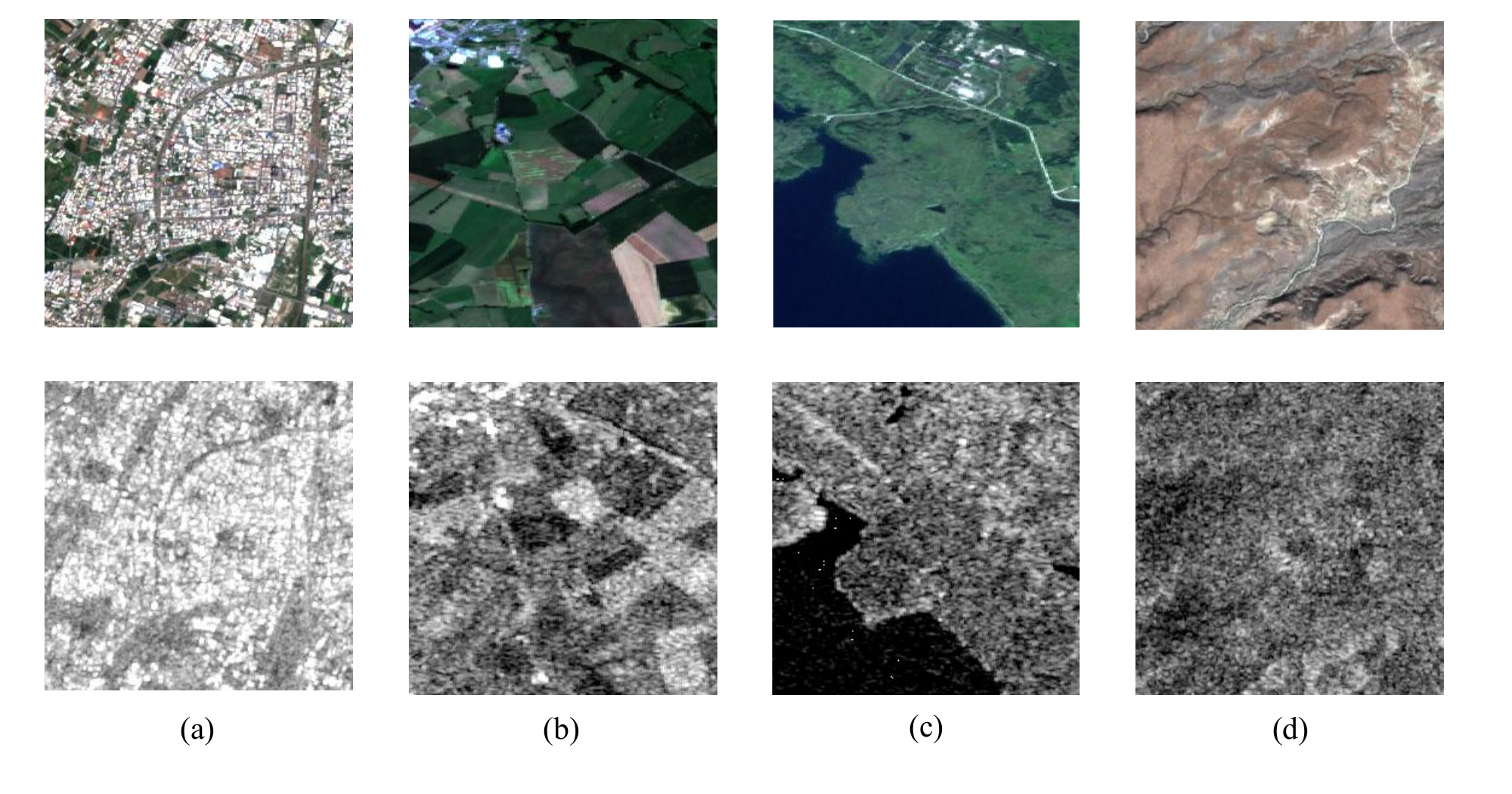}
  \caption{\textbf{Cross-modal sensor comparison.}
    Sentinel-2 multispectral (top) and Sentinel-1 SAR (bottom) across
    four land-cover types: (a)~Urban, (b)~Farmland, (c)~Grassland,
    (d)~Barren~\cite{tiwari2020sentinel}.
    The gap between passive optical reflectance and active microwave backscatter
    is visually apparent across all categories.}
  \label{fig:sensor_comp}
\end{figure}

Synthetic aperture radar (SAR) provides all-weather, day-and-night imaging critical for environmental
monitoring, disaster response, and defense
intelligence~\cite{6504845,8113128,8291019}.
Unlike optical imagery, SAR backscatter encodes coherent scattering speckle and
viewing geometry with no optical analog (Fig.~\ref{fig:sensor_comp}) ~\cite{chen2025purelyspeckledintensityimages,9884596},
and data scarcity compounds the challenge. Generating accurate radar captions
requires deep domain expertise, and even the largest SAR caption
datasets~\cite{Wei2025SARLANG1M,He2025SARTEXT,Jiang2026SARCLIP,11142698}
are orders of magnitude smaller than the optical corpora used to pre-train vision-language models. As a result, adapting such models to SAR imagery involves two compounding failure modes that current
methods address inadequately, and which motivate the design of our approach.
\begin{figure*}[t]
  \centering
  \includegraphics[width=0.88\linewidth]{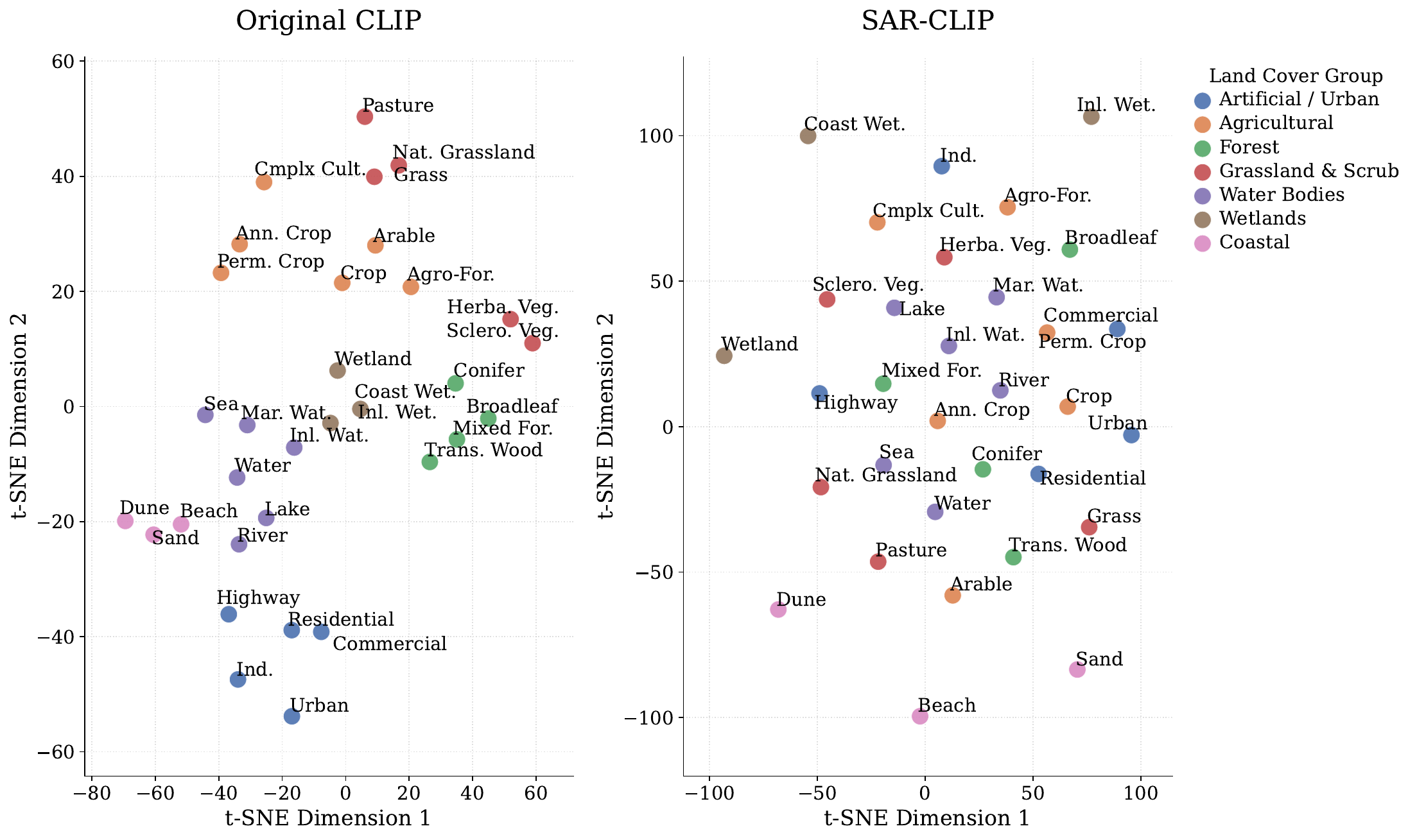}
  \caption{\textbf{Manifold distortion under full co-training.}
    $t$-SNE of the text encoder embedding space for pretrained CLIP (left) vs.\
    fully fine-tuned SAR-CLIP~\cite{Jiang2026SARCLIP} (right).
    Joint fine-tuning on SAR data induces geometric warping and inter-class
    entanglement across the semantic clusters, consistent with impaired
    open-vocabulary zero-shot generalization.}
  \label{fig:manifold}
\end{figure*}
The first failure mode is \emph{modality mismatch}.
Text encoders such as CLIP are pre-trained on hundreds of millions of optical
image-text pairs, where the visual signal is passive reflectance of light from surfaces.
The physics of radar backscatter has no analog in this training regime:
SAR encodes active microwave scattering mechanisms (double-bounce, volumetric diffuse,
surface roughness) that simply do not appear in optical imagery.
Fine-tuning CLIP on SAR data pushes the model to adapt to a physics regime
on datasets that are orders of magnitude smaller than the original 400M-pair
optical pre-training corpus.
The result is a predictable trade-off: the broad open-vocabulary coverage that makes CLIP useful
is substantially compressed by narrow, synthetically generated caption distributions,
and the semantic manifold is geometrically warped in the process (Fig.~\ref{fig:manifold}).

The second failure mode is \emph{physical forgetting}.
SAR foundation models such as CROMA~\cite{NEURIPS2023_11822e84} encode
radar-specific physics through large-scale self-supervised pre-training:
cross-modal contrastive learning against optical imagery teaches the encoder
that specific backscatter textures correspond to recognizable optical
structures~\cite{6504845,9884596}.
These physically grounded representations are the SAR encoder's most valuable asset
for downstream alignment.
Yet when the encoder is fine-tuned to minimize a cross-modal contrastive loss against
text embeddings on a small SAR caption dataset, gradient updates optimize for
text proximity at the expense of physical fidelity.
Our ablations (Table~\ref{tab:ablation_freezing}) and Procrustes analysis
(Table~\ref{tab:procrustes}) show that this distortion is measurable and substantial.

We propose a Frozen Expert framework.
Rather than retraining either encoder, we keep both strictly frozen and learn only
a constrained, structure-preserving bridge between them.
The frozen CROMA encoder provides physically grounded SAR representations.
The frozen CLIP text encoder provides a stable open-vocabulary semantic space.
Since CROMA is grounded in optical imagery, the alignment follows a transitive chain
(SAR $\to$ Optical $\to$ Text), reducing reliance on noisy radar captions
for semantic grounding. Despite training on only $\sim$50{,}000 image-caption pairs compared to
130{,}000+ used by SAR-RS-CLIP~\cite{He2025SARTEXT} and
430{,}000+ used by SARCLIP~\cite{Jiang2026SARCLIP}, our model consistently
outperforms both across DFC2020~\cite{9369830}, EuroSAT-SAR~\cite{Wang2024EuroSATSAR},
and BigEarthNet-MM~\cite{9552024}.

\noindent Our contributions are:
\begin{itemize}
  \item[--] \textbf{Principle.} We formally characterize the two failure modes
        of high-disparity cross-modal adaptation and show that freezing both
        encoders substantially mitigates both simultaneously.
        We define physical forgetting via Procrustes disparity and demonstrate
        empirically that our bridge preserves SAR geometry better than fine-tuning.
  \item[--] \textbf{Method.} A Frozen Expert framework with a fully formalized
        constrained structure-preserving bridge
        (Eqs.~\ref{eq:bridge}--\ref{eq:infonce}) that maps frozen SAR
        representations into the frozen CLIP text manifold without gradient
        updates to either encoder.
  \item[--] \textbf{Generality.} We validate across multiple SAR backbones and
        text encoders, confirming that the frozen-alignment principle drives
        the gains, not the specific choice of CROMA or CLIP.
  \item[--] \textbf{State of the art.} New benchmarks for zero-shot SAR scene
        classification across three datasets at $8\times$ less training data
        than leading fine-tuned baselines.
\end{itemize}

\section{Related Work}
\label{sec:related}

\subsection{VLMs for Remote Sensing}
The success of CLIP~\cite{Radford2021CLIP} spurred a wave of optical remote sensing
adaptations, including RemoteCLIP~\cite{10504785} and GeoRSCLIP~\cite{10679571},
which fine-tune CLIP on large optical captioning datasets.
These models fail to generalize to SAR due to the profound physical domain gap.
The community has pivoted toward SAR-specific VLMs, fueled by the curation of
large-scale SAR caption
datasets~\cite{Wei2025SARLANG1M,He2025SARTEXT,Jiang2026SARCLIP,11142698}.
We argue that scale mismatch renders these approaches fragile: even the largest
SAR caption datasets are orders of magnitude smaller than the 400M optical pairs
used to pre-train CLIP, making fine-tuning prone to catastrophic
forgetting~\cite{Vesdapunt2024HVCLIP,Kumar2022FineTuning}.
Moreover, adapting to SAR is categorically different from adapting between
optical domains: radar backscatter encodes double-bounce scattering, volumetric
diffuse returns, and surface roughness effects that have no direct optical
analog~\cite{6504845,9884596}, and optical pre-training alone does not readily
endow an encoder with these physical concepts.

\subsection{Self-Supervised Learning in SAR}
Contrastive methods~\cite{9880533,9614157} and reconstruction-based
approaches~\cite{Wang2025MIM,10766851} have been adapted for SAR, with recent
transformer architectures~\cite{sarformer2024} extending SAR-specific
representation learning.
The most powerful representations emerge from multimodal
pre-training~\cite{Astruc2024OmniSat,NEURIPS2023_11822e84}.
CROMA~\cite{NEURIPS2023_11822e84} employs cross-modal contrastive learning
between SAR and optical imagery alongside masked autoencoding, grounding its SAR
encoder in visual physics.
Feature Guided MAE (FG-MAE)~\cite{10766851}, pre-trained on the same SSL4EO
dataset as CROMA but using only the SAR channel without an optical anchor,
provides a near-controlled baseline in our backbone ablation
(Sec.~\ref{sec:ablations}, Table~\ref{tab:backbone_generality}):
same data, same scale, same modality, but without cross-modal grounding.
We additionally include a JEPA-pretrained SAR encoder~\cite{sarjepa}
in the backbone ablation to assess a model pre-trained on SAR target-recognition
data, which lacks scene-level land-cover priors.

\subsection{VLM Adaptation Strategies}
\label{sec:relwork_adaptation}

As vision-language foundation models grow, retraining them from scratch becomes
computationally prohibitive.
Parameter-efficient fine-tuning methods, including low-rank adaptation and adapter
layers, have been proposed to reduce the cost of
adaptation~\cite{Zanella2024LowRankFewShotVLM,Zhang2025AdaptiveParameterSelection,Wang2025LoRALoop,Li2025NestedLoRA,Mai2025PEFTLessons},
but even these constrained approaches can induce manifold distortion when applied
to non-optical
domains~\cite{pmlr-v202-shen23e,NEURIPS2023_a45296e8,agarwal2026liteembedadaptingcliprare,chen2026preserve,rajaee2021how,ypsilantis2025infusing}.

Merullo et al.~\cite{merullo2023linearly} demonstrated that a linear map can align
optical vision representations with language, exploiting the shared semantic topology
between optical imagery and text: visual attributes like shape and color have
direct linguistic equivalents, making the alignment geometrically tractable.
In the SAR domain, this semantic overlap is substantially weaker.
Radar backscatter encodes physical scattering mechanisms with no natural-language
analog, so a linear map lacks the expressivity to navigate the topological
difference between SAR and text manifolds.
Our linear projection ablation confirms this, as we discuss further in
Sec.~\ref{sec:discussion}.

Locked-Image Tuning (LiT)~\cite{9878889} showed that freezing a strong image
encoder and training only the text encoder yields competitive zero-shot transfer,
indicating that high-quality visual representations do not need to be re-trained
for alignment.
Our work extends this insight in two directions.
First, we freeze the text encoder as well as the image encoder: SAR captions
lack the semantic breadth to regularize CLIP's manifold, and even RS-specialized
text encoders degrade open-vocabulary performance relative to the original CLIP,
as we demonstrate quantitatively in Sec.~\ref{sec:discussion}.
Second, we freeze the image encoder not merely because it is already good at its
task, but because it encodes physical priors that are actively destroyed by
gradient-based text alignment on limited data.

Swift Alignment~\cite{11093598} explores alignment of unimodal vision and language
models for natural images.
Our work shares the philosophy but addresses a fundamentally harder regime: the signal
formation process changes at the sensor, not just the semantic distribution.
Our findings are consistent with, and may further motivate, the Platonic
Representation Hypothesis~\cite{pmlr-v235-huh24a}: convergent cross-modal
pre-training may induce compatible internal geometries that can be bridged without
retraining, even across fundamentally different physical modalities.

\section{Methodology}
\label{sec:method}

\begin{figure*}[t]
  \centering
  \includegraphics[width=0.88\linewidth]{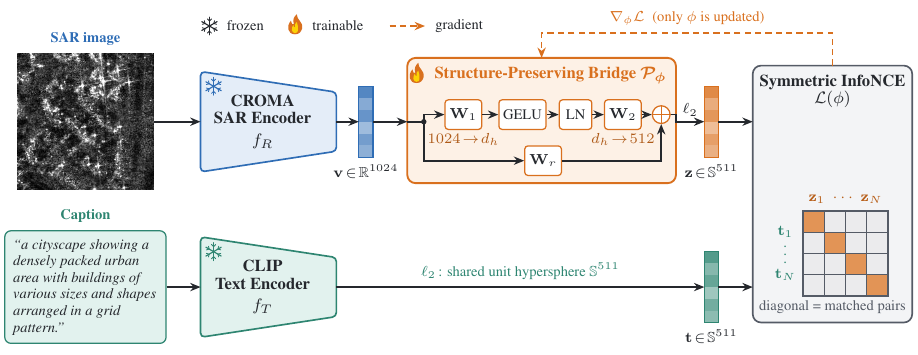}
  \caption{\textbf{Overview of our Frozen Expert framework.}
    A constrained structure-preserving bridge $\mathcal{P}_\phi$ maps frozen CROMA
    SAR representations into the frozen CLIP text manifold.
    The bridge (Eq.~\ref{eq:bridge}) combines a two-layer MLP with a linear residual
    path, and its $\ell_2$-normalized output $\bz$ is aligned with the caption embedding
    $\bt$ on the shared unit hypersphere via a symmetric InfoNCE loss (Eq.~\ref{eq:infonce}).
    Only $\phi$ receives gradient updates; both encoders remain parameter-frozen
    throughout training.}
  \label{fig:arc}
\end{figure*}

Our objective is to align SAR imagery with natural language without massive
annotated datasets.
Our Frozen Expert framework (Fig.~\ref{fig:arc}) uses two specialized, frozen
networks: a physically grounded SAR encoder (Sec.~\ref{sec:visual_backbone}) and
the CLIP text encoder (Sec.~\ref{sec:semantic_anchor}), connected by a constrained
structure-preserving bridge (Sec.~\ref{sec:bridge}).

\subsection{Visual Backbone: Physically Grounded Frozen Expert}
\label{sec:visual_backbone}

We use the radar encoder $f_R$ from CROMA~\cite{NEURIPS2023_11822e84} as our
visual backbone, chosen because its cross-modal pre-training creates a physics-grounded
manifold uniquely suited to semantic alignment.

\noindent\textbf{Architecture.}
$f_R$ is a Vision Transformer processing dual-polarized Sentinel-1 data via 2D
attention with linear biases over $8\!\times\!8$ non-overlapping patches,
preserving geometric relationships across diverse remote sensing resolutions.

\noindent\textbf{Pre-training on SSL4EO.}
CROMA is pre-trained on SSL4EO-S12~\cite{10261879}, a large-scale dataset
containing over 1 million multi-temporal SAR-optical paired observations across
250K global locations, each captured at four seasonal snapshots.
Its dual objective, cross-modal contrastive learning between SAR and optical
plus masked autoencoding, forces the encoder to implicitly model the visual
structure of scenes: specific backscatter textures are associated with corresponding
optical structures (e.g., forest canopy roughness, water specular reflection).

\noindent\textbf{Manifold preservation.}
During alignment, $\theta_R$ is strictly frozen (inference mode, no batch-norm
updates).
A global image representation $\bv \in \Rset^{D_v}$ ($D_v = 1024$) is produced
by global average pooling over patch embeddings followed by the backbone's
pre-trained projection head.
Freezing $\theta_R$ anchors $\bv$ in the robust, sensor-invariant physics
distilled from SSL4EO.

\subsection{Semantic Anchor: CLIP Embedding Space}
\label{sec:semantic_anchor}

We employ the CLIP ViT-B/32 text encoder~\cite{Radford2021CLIP} as a fixed
geometric anchor for two reasons.
First, CLIP provides a coherent open-vocabulary vector space enabling zero-shot
inference over arbitrary class names.
Second, available SAR captions are synthetically generated and noisy; fine-tuning
on such data can substantially degrade the robust semantic structure of the language model,
as evidenced by our text encoder swap ablation (Table~\ref{tab:encoder_generality}).

Text caption $T$ is tokenized (max 48 tokens) and encoded through $f_T$
(parameters $\theta_T$, frozen, dropout disabled):
\begin{equation}
  \bt = \frac{f_T(T)}{\|f_T(T)\|_2} \in \mathbb{S}^{d_t - 1}, \quad d_t = 512.
  \label{eq:text_enc}
\end{equation}
Here $\mathbb{S}^{d_t-1} = \{x \in \Rset^{d_t} : \|x\|_2 = 1\}$ is the unit
hypersphere embedded in $\Rset^{d_t}$: L2-normalization constrains the output to
this surface so that cosine similarity equals the inner product.
Both the text embedding $\bt$ and the bridge output $\bz$ (Eq.~\ref{eq:bridge})
are constrained to lie on this surface.
Because $\theta_T$ is frozen, pairwise semantic distances (e.g., the cosine
similarity between ``dense forest'' and ``rough canopy'') remain constant,
compelling the bridge to conform to this stable organization.

\subsection{The Bridge: Constrained Structure-Preserving Alignment}
\label{sec:bridge}

\noindent\textbf{Architecture.}
The bridge $\mathcal{P}_\phi: \Rset^{D_v} \to \mathbb{S}^{d_t-1}$ maps frozen
SAR representations into the text manifold:
\begin{equation}
  \mathcal{P}_\phi(\bv) = \ell_2\!\left(
    \bW_2\,\LN\!\left(\GELU(\bW_1\bv + \bb_1)\right) + \bb_2 + \bW_r\bv
  \right),
  \label{eq:bridge}
\end{equation}
where $\bW_1 \!\in\! \Rset^{d_h \times D_v}$,
      $\bW_2 \!\in\! \Rset^{d_t \times d_h}$,
      $\bW_r \!\in\! \Rset^{d_t \times D_v}$
are learnable parameters, $\LN$ is Layer Normalization, $\ell_2(\cdot)$ is
$\ell_2$-normalization, and $d_h$ is the hidden dimension ablated in
Sec.~\ref{sec:ablations}.
The residual term $\bW_r\bv$ preserves direct signal flow across the
dimension reduction ($D_v\!=\!1024 \to d_t\!=\!512$).

\noindent\textbf{``Constrained'' defined.}
We use ``constrained'' to denote the deliberate capacity restriction:
$|\phi| \ll |\theta_R|, |\theta_T|$.
A moderate hidden dimension prevents the bridge from distorting the input physical
geometry to force-fit caption noise; excessive capacity actively degrades
performance (Table~\ref{tab:ablation_head}).

\noindent\textbf{Structure preservation and physical forgetting.}
The dimension reduction ($D_v \to d_t$) and non-linearities preclude strict
topological isomorphism.
We instead formalize an empirically grounded notion of structure preservation
using normalized Procrustes disparity~\cite{gower2004procrustes}:
\begin{equation}
  \Delta(\bA, \bB) = \min_{\mathbf{R}\in\mathcal{O}(d),\,c>0}\;
    \frac{\|\hat{\bA} - c\hat{\bB}\mathbf{R}^\top\|_F}{\|\hat{\bA}\|_F},
  \label{eq:procrustes}
\end{equation}
where $\hat{(\cdot)}$ denotes row-centering, $\mathbf{R}$ is an orthogonal matrix,
and $c$ is a scale factor.
Lower $\Delta$ indicates that the output geometry mirrors the input after optimal
rigid alignment.
We define physical forgetting as
$\Delta(\mathbf{V}_\text{frozen}, \mathbf{V}_\text{tuned})$: the Procrustes
disparity between frozen and fine-tuned CROMA representations on the same inputs.
This operationalizes the intuition that gradient updates on limited SAR captions
rearrange the scattering-geometry structure encoded by the pre-trained encoder.

Table~\ref{tab:procrustes} validates this empirically on DFC2020 test images:
our frozen bridge (row A, $\Delta\!=\!0.210$) preserves CROMA geometry better
than fine-tuning preserves CROMA's own manifold (row B, $\Delta\!=\!0.351$;
$+67\%$ distortion).
End-to-end unfreezing (row C, $\Delta\!=\!0.388$) produces the worst distortion.

\noindent\textbf{Training objective.}
Only $\phi$ receives gradients; $\theta_R$ and $\theta_T$ remain frozen.
We minimize the symmetric bidirectional InfoNCE loss:
\begin{equation}
  \mathcal{L}(\phi) = -\frac{1}{2N}\sum_{i=1}^{N}\!\left[
    \log\frac{e^{\bz_i^\top \bt_i/\tau}}{\sum_j e^{\bz_i^\top \bt_j/\tau}}
    + \log\frac{e^{\bt_i^\top \bz_i/\tau}}{\sum_j e^{\bt_i^\top \bz_j/\tau}}
  \right],
  \label{eq:infonce}
\end{equation}
where $\bz_i = \mathcal{P}_\phi(\bv_i)$ and $\tau = 0.07$ is fixed.

\section{Experimental Setup}
\label{sec:setup}

\subsection{Training Dataset}
We restrict training to a $\sim\!50{,}000$ image-caption subset of
SAR-TEXT~\cite{He2025SARTEXT} (SEN12, WHU-SAR-OPT, QXSLAB-SAROPT, OSdataset
collections).
This is intentionally low-resource: baselines SAR-RS-CLIP and SARCLIP require
130{,}000+ and 430{,}000+ pairs, respectively, to achieve competitive performance.

\subsection{Zero-Shot Evaluation Datasets}
\textbf{DFC2020}~\cite{9369830}: A multi-label challenge using $96\!\times\!96$
SAR patches across 8 land-cover classes.
Each patch contains multiple ground-truth labels reflecting the heterogeneity of
complex scenes.

\noindent \textbf{EuroSAT-SAR}~\cite{Wang2024EuroSATSAR}: 27{,}000 single-label $64\!\times\!64$ SAR
patches across 10 land-cover classes.
It requires fine discrimination between visually similar categories
(e.g., ``Annual Crop'' vs.\ ``Permanent Crop'').

\noindent \textbf{BigEarthNet-MM}~\cite{9552024}: A 19-class multi-label dataset using
$120\!\times\!120$ patches.
The fine-grained taxonomy requires distinguishing between semantically overlapping
concepts without prior fine-tuning.

 These benchmarks vary in task complexity (single- vs.\ multi-label),
resolution, class granularity, and geographic coverage,
providing a rigorous test of zero-shot generalization.
\begin{table}[t]
  \centering
  \caption{\textbf{Procrustes disparity validates structure preservation.}
    $\Delta\!\downarrow$ measures normalized geometric distortion relative to
    frozen CROMA representations.
    Our bridge distorts SAR geometry less than fine-tuning distorts CROMA's own
    manifold.}
  \label{tab:procrustes}
  \small
  \begin{tabular}{clc}
    \toprule
     & \textbf{Comparison} & $\Delta\!\downarrow$ \\
    \midrule
    A & CROMA\textsubscript{frozen} $\leftrightarrow$ Bridge (Ours) & \textbf{0.210} \\
    B & CROMA\textsubscript{frozen} $\leftrightarrow$ CROMA\textsubscript{fine-tuned} & 0.351 \\
    C & CROMA\textsubscript{frozen} $\leftrightarrow$ Bridge\textsubscript{end-to-end} & 0.388 \\
    \bottomrule
  \end{tabular}
\end{table}
\subsection{Baseline Models}
\textbf{SAR-RS-CLIP}~\cite{He2025SARTEXT}: A ViT-L/14 model applying two-stage
progressive transfer from an optical foundation model, fine-tuned on
$\sim\!130{,}000$ SAR-TEXT pairs via full-model contrastive training.

\noindent \textbf{SARCLIP}~\cite{Jiang2026SARCLIP}: A ViT-L/14 framework trained on
$\sim\!430{,}000$ rule-based SARCAP pairs, augmented with a Noise-Robust Encoding
module and Hierarchical Prompt Learning for multi-scale semantic alignment.

To our knowledge, SARCLIP and SAR-RS-CLIP are the only published vision-language models that support open-vocabulary zero-shot SAR classification under the same evaluation protocol. Existing PEFT methods such as LoRA, CoOp, CLIP-Adapter, and prompt tuning are generic adaptation techniques rather than complete SAR vision-language systems. Constructing SAR-specific variants would require introducing new architectures and training protocols, yielding author-implemented baselines rather than published methods. We therefore compare against the established state of the art for SAR zero-shot vision-language modeling.

\subsection{Metrics}

\noindent\textbf{Recall@K (R@K).}
We adapt R@K to assess the model's ability to prioritize relevant semantic concepts.
For the multi-label datasets (DFC2020, BigEarthNet-MM), R@K measures the percentage
of test images where at least one correct ground-truth label appears within the
top-$K$ predicted classes.
For the single-label EuroSAT-SAR dataset, this simplifies to standard top-$K$
accuracy, measuring whether the single ground-truth class is among the top-$K$
predictions.
We report this for $K \in \{1, 2, 3\}$.

\begin{table*}[h]
  \centering
  \caption{\textbf{Effect of projection head capacity (DFC2020).}
    All models use frozen CROMA and frozen CLIP. Bold: best per column.}
  \label{tab:ablation_head}
  \resizebox{0.92\linewidth}{!}{%
  \begin{tabular}{l c | cccccccc | c}
    \toprule
    \textbf{Head Type} & $d_h$ &
    \textbf{R@1} & \textbf{R@2} & \textbf{R@3} & \textbf{R@4} &
    \textbf{R@5} & \textbf{R@6} & \textbf{R@7} & \textbf{R@8} & \textbf{mAP} \\
    \midrule
    Linear & N/A  & 23.14 & 30.34 & 64.44 & 85.75 & 93.10 & 96.82 & 98.22 & 100.00 & 15.91 \\
    MLP    & 256  & 59.92 & 72.15 & 82.03 & 89.98 & 95.20 & 96.90 & 98.66 & 100.00 & 33.57 \\
    MLP    & 512  & \textbf{66.30} & \textbf{79.41} & \textbf{88.28} & \textbf{93.56} & \textbf{97.42} & 98.73 & 99.75 & 100.00 & \textbf{36.56} \\
    MLP    & 1024 & 64.04 & 75.54 & 84.75 & 89.52 & 96.06 & \textbf{99.24} & \textbf{99.90} & 100.00 & 29.61 \\
    MLP    & 2048 & 44.00 & 65.19 & 76.19 & 81.54 & 85.23 & 95.26 & 98.74 & 100.00 & 23.49 \\
    \bottomrule
  \end{tabular}}
\end{table*}

\begin{table*}[h]
  \centering
  \caption{\textbf{Impact of expert selection and freezing strategy (DFC2020).}
    Bold: best per column.
    Domain-specific SAR expertise and strict encoder freezing are both essential.}
  \label{tab:ablation_freezing}
  \resizebox{0.92\linewidth}{!}{%
  \begin{tabular}{l l l | cccccccc | c}
    \toprule
    \textbf{Visual Enc.} & \textbf{Text Enc.} & \textbf{Strategy} &
    \textbf{R@1} & \textbf{R@2} & \textbf{R@3} & \textbf{R@4} &
    \textbf{R@5} & \textbf{R@6} & \textbf{R@7} & \textbf{R@8} & \textbf{mAP} \\
    \midrule
    CLIP (Optical) & CLIP & Frozen       & 34.41 & 54.60 & 70.40 & 82.46 & 90.30 & 92.95 & 96.10 & 100.00 & 17.71 \\
    CROMA & CLIP & Unfrozen Text         & 10.35 & 16.40 & 22.79 & 65.79 & 82.51 & 89.33 & 93.10 & 100.00 & 15.29 \\
    CROMA & CLIP & Unfrozen Image        & 52.69 & 54.66 & 60.08 & 65.94 & 74.67 & 81.11 & 89.82 & 100.00 & 15.27 \\
    CROMA & CLIP & \textbf{Frozen (Ours)}& \textbf{66.30} & \textbf{79.41} & \textbf{88.28} & \textbf{93.56} & \textbf{97.42} & \textbf{98.73} & \textbf{99.75} & 100.00 & \textbf{36.56} \\
    \bottomrule
  \end{tabular}}
\end{table*}

\noindent\textbf{Mean Average Precision (mAP).}
We report mAP as an additional performance indicator for all datasets.
For each class $c$, we compute the Average Precision (AP) from the
precision-recall curve derived from the model's confidence scores.
The mAP is calculated as the mean over all classes. During inference, we compute the cosine similarity between the $\ell_2$-normalized
image embedding $\bv$ and the frozen text embeddings $\bt_c$ for all candidate
classes $c \in \mathcal{C}$, properly scaled with respect to temperature.
To obtain class probabilities, we apply activation functions based on the task
nature.
For the single-label EuroSAT-SAR dataset, we apply a softmax activation to produce
a categorical probability distribution, enforcing a single class prediction per image.
For the multi-label DFC2020 and BigEarthNet-MM datasets, we apply a sigmoid
activation to generate independent probabilities for each class, as multiple
classes may coexist in a single patch.


\section{Results}
\label{sec:results}

\subsection{Ablations}
\label{sec:ablations}

All ablations use DFC2020 as the primary evaluation dataset.

\noindent\textbf{Projection head capacity (Table~\ref{tab:ablation_head}).}
We compare a linear projection against 2-layer MLPs with hidden dimensions
$d_h \in \{256, 512, 1024, 2048\}$.
The residual MLP consistently outperforms the linear baseline, validating the need
for a non-linear mapping to bridge SAR and text manifolds.
Performance peaks at $d_h\!=\!512$ ($36.56\%$ mAP, $66.30\%$ R@1).
Increasing capacity to $d_h\!=\!2048$ degrades significantly, confirming that
excessive capacity allows the bridge to distort SAR geometry to fit caption noise
rather than finding a genuine geometric alignment.
``Constrained'' capacity is therefore a necessary regularizer.
We fix $d_h\!=\!512$ for all subsequent experiments.

\noindent\textbf{Manifold preservation strategy (Table~\ref{tab:ablation_freezing}).}
Holding $d_h\!=\!512$ fixed, we evaluate our Frozen Expert strategy against alternatives.
Replacing CROMA with the frozen optical CLIP image encoder drops R@1 by $31.9\%$
($66.30\% \to 34.41\%$), confirming that optical pre-training fails to interpret
SAR geometry and scattering statistics.
Unfreezing the text encoder causes catastrophic collapse (R@1: $10.35\%$),
reflecting that limited SAR captions corrupt the CLIP semantic manifold.
Unfreezing CROMA significantly degrades mAP ($36.56 \to 15.27$), consistent with
physical forgetting: gradient updates on caption data displace the
physics-grounded representations learned during self-supervised pre-training.

\noindent\textbf{Text encoder generality (Table~\ref{tab:encoder_generality}).}
We replace the frozen CLIP text encoder with RS-specialized encoders from
GeoRSCLIP~\cite{10679571} and RemoteCLIP~\cite{10504785},
retraining the bridge from scratch on the same $\sim\!50$K pairs.
Performance degrades substantially with each swap ($66.30 \to 52.97 \to 38.82$
R@1), confirming that RS-domain fine-tuning warps the global semantic distances
required for open-vocabulary zero-shot transfer.
The original CLIP text manifold, despite never having encountered SAR imagery,
provides the most suitable semantic anchor.

\noindent\textbf{SAR backbone generality (Table~\ref{tab:backbone_generality}).}
We swap the frozen SAR encoder while retraining only the bridge from scratch.
FG-MAE~\cite{10766851} is pre-trained on SSL4EO-S12 at the same scale as CROMA
but uses only the SAR channel, ignoring paired optical imagery entirely.
The $\sim\!12\%$ R@1 gap ($66.30 \to 54.06$) directly isolates the benefit of
cross-modal optical grounding: same data distribution, same scale, same modality;
only the presence of the optical anchor differs.
Using the SAR-RS-CLIP image encoder ($63.17\%$ R@1) surpasses SAR-RS-CLIP's own
zero-shot performance ($55.90\%$, Table~\ref{tab:merged_results}), confirming
that the frozen bridge generalizes across backbones and the bottleneck lies in
the alignment strategy, not necessarily the visual backbone.
The JEPA-pretrained SAR encoder~\cite{sarjepa} yields near-chance retrieval, as it
was pre-trained on SAR target-recognition data and lacks scene-level land-cover
priors or optical grounding.
\begin{table}[t]
  \centering
  \caption{\textbf{Text encoder generality (DFC2020).}
    Bridge retrained from scratch for each encoder.
    RS fine-tuning degrades open-vocabulary transfer.}
  \label{tab:encoder_generality}
  \small
  \begin{tabular}{l cccc}
    \toprule
    \textbf{Text Encoder} & \textbf{R@1} & \textbf{R@2} & \textbf{R@3} & \textbf{mAP} \\
    \midrule
    \textbf{CLIP (Ours)} & \textbf{66.30} & \textbf{79.41} & \textbf{88.28} & \textbf{36.56} \\
    GeoRSCLIP~\cite{10679571} & 52.97 & 64.92 & 76.72 & 31.19 \\
    RemoteCLIP~\cite{10504785} & 38.82 & 65.82 & 80.09 & 27.45 \\
    \bottomrule
  \end{tabular}
\end{table}
\subsection{Zero-Shot Performance}
\label{sec:zeroshot}
\begin{table}[t]
  \centering
  \caption{\textbf{SAR backbone generality (DFC2020).}
    Bridge retrained from scratch for each backbone;
    CLIP text encoder frozen throughout.
    Cross-modal grounding drives performance, not backbone scale.}
  \label{tab:backbone_generality}
  \small
  \begin{tabular}{l cc}
    \toprule
    \textbf{SAR Backbone} & \textbf{R@1} & \textbf{mAP} \\
    \midrule
    \textbf{CROMA~\cite{NEURIPS2023_11822e84} (Ours)} & \textbf{66.30} & \textbf{36.56} \\
    SAR-RS-CLIP enc.~\cite{He2025SARTEXT}             & 63.17 & 21.42 \\
    FG-MAE~\cite{10766851} (SAR-only SSL)              & 54.06 & 24.50 \\
    JEPA-SAR enc.~\cite{sarjepa} (target recog.)       & 13.84 & 16.35 \\
    \bottomrule
  \end{tabular}
\end{table}
Table~\ref{tab:merged_results} compares our Frozen Expert framework against
SAR-RS-CLIP and SARCLIP across all three benchmarks.
We report our model at the two best-performing projection dimensions from the
capacity ablation ($d_h\!=\!512$ and $d_h\!=\!1024$).

\noindent\textbf{Recall.}
Our model achieves $66.30\%$ R@1 on DFC2020, surpassing SARCLIP ($56.01\%$) and
SAR-RS-CLIP ($55.90\%$) by over 10 absolute percentage points.
The gap is particularly striking given that both baselines use ViT-L/14 backbones
while we use ViT-B/32, suggesting that the alignment strategy, not backbone
capacity, is the binding constraint.
On EuroSAT-SAR, we nearly double both baselines: $33.44\%$ vs.\ $17.79\%$ and
$14.72\%$.
This dataset's single-label classification task, where only one correct class
exists per image, is particularly sensitive to manifold distortion: a warped
text embedding space produces incorrect similarity rankings even when the image
embedding is valid.
On BigEarthNet-MM, our model achieves $33.05\%$ R@1 vs.\ $18.34\%$ (SARCLIP) and
$17.64\%$ (SAR-RS-CLIP), consistent superiority across datasets that vary
substantially in task complexity, resolution, and class granularity.

\noindent\textbf{Mean Average Precision.}
On DFC2020, we achieve $36.56\%$ mAP, outperforming SARCLIP ($21.05\%$) by
$+15.5$ points.
On EuroSAT-SAR, we reach $35.11\%$ vs.\ $22.87\%$.
On BigEarthNet-MM, while SARCLIP achieves a marginally higher mAP ($21.44\%$)
than our model ($20.97\%$), we still significantly outperform SAR-RS-CLIP ($16.31\%$).
Additionally, it is crucial to highlight the label efficiency: SARCLIP was trained
on $\sim\!430{,}000$ pairs, whereas our model was aligned using only
$\sim\!50{,}000$ pairs.
Achieving near-parity in mAP while dominating in early retrieval (including an
absolute $+14.71\%$ lead in R@1) with an $8\times$ reduction in training data
highlights the extreme efficiency of bridging frozen experts.
This indicates that massive, noisy data regimes yield diminishing returns for
global precision, whereas our approach excels at immediately prioritizing the
most relevant semantic matches.

\begin{table}[t]
  \centering
  \caption{\textbf{Zero-shot classification performance.}
    Our framework vs.\ SAR-RS-CLIP and SARCLIP across three benchmarks.
    Bold: best per dataset per metric.}
  \label{tab:merged_results}
  \resizebox{\columnwidth}{!}{%
  \begin{tabular}{l l | cccc}
    \toprule
    \textbf{Dataset} & \textbf{Model} &
    \textbf{R@1} & \textbf{R@2} & \textbf{R@3} & \textbf{mAP} \\
    \midrule
    \multirow{4}{*}{DFC2020}
     & SAR-RS-CLIP~\cite{He2025SARTEXT}  & 55.90\% & 58.94\% & 74.35\% & 17.29\% \\
     & SARCLIP~\cite{Jiang2026SARCLIP}   & 56.01\% & 67.37\% & 87.82\% & 21.05\% \\
     & Ours ($d_h\!=\!512$) & \textbf{66.30\%} & \textbf{79.41\%} & \textbf{88.28\%} & \textbf{36.56\%} \\
     & Ours ($d_h\!=\!1024$)& 64.04\% & 75.54\% & 84.75\% & 29.61\% \\
    \midrule
    \multirow{4}{*}{EuroSAT-SAR}
     & SAR-RS-CLIP~\cite{He2025SARTEXT}  & 14.72\% & 25.77\% & 38.28\% & 22.87\% \\
     & SARCLIP~\cite{Jiang2026SARCLIP}   & 17.79\% & 32.80\% & 43.28\% & 22.05\% \\
     & Ours ($d_h\!=\!512$) & 31.72\% & 51.90\% & 64.20\% & 34.82\% \\
     & Ours ($d_h\!=\!1024$)& \textbf{33.44\%} & \textbf{52.04\%} & \textbf{65.95\%} & \textbf{35.11\%} \\
    \midrule
    \multirow{4}{*}{BigEarthNet}
     & SAR-RS-CLIP~\cite{He2025SARTEXT}  & 17.64\% & 42.43\% & 47.24\% & 16.31\% \\
     & SARCLIP~\cite{Jiang2026SARCLIP}   & 18.34\% & 32.72\% & 47.52\% & \textbf{21.44\%} \\
     & Ours ($d_h\!=\!512$) & \textbf{33.05\%} & \textbf{59.00\%} & \textbf{71.64\%} & 20.97\% \\
     & Ours ($d_h\!=\!1024$)& 30.27\% & 52.86\% & 63.83\% & 20.71\% \\
    \bottomrule
  \end{tabular}}
\end{table}
\section{Discussion}
\label{sec:discussion}
 
\noindent\textbf{Alignment strategy comparisons.}
Our ablation results directly ground the distinctions from prior alignment
strategies drawn in Sec.~\ref{sec:relwork_adaptation}.
For Merullo et al.~\cite{merullo2023linearly}: a linear bridge achieves only
$23.1\%$ R@1 on DFC2020 compared to $66.3\%$ for our residual MLP
(Table~\ref{tab:ablation_head}), a 43-point gap that confirms radar backscatter
lacks the semantic-optical overlap that makes linear alignment tractable in the
optical domain.
The non-linearity in our bridge is not an arbitrary design preference but a
necessity imposed by the topological difference between SAR and text manifolds.
For LiT~\cite{9878889}: unfreezing the text encoder alone collapses R@1 from
$66.3\%$ to $10.3\%$ (Table~\ref{tab:ablation_freezing}), and the severity of
degradation across RS-specialized text encoders scales inversely with their
fine-tuning data volume: GeoRSCLIP, fine-tuned on $\sim$5M optical RS pairs,
retains substantially more open-vocabulary coverage ($52.97\%$ R@1) than
RemoteCLIP ($\sim$830K pairs, $38.82\%$ R@1), yet both degrade significantly
relative to original CLIP ($66.30\%$ R@1, trained on 400M pairs;
Table~\ref{tab:encoder_generality}).
 
\noindent\textbf{Both failure modes, one design principle.}
Our results validate that the two failure modes identified in the introduction
are real and that a single design principle substantially addresses both.
On the text encoder side, Table~\ref{tab:ablation_freezing} shows that unfreezing
CLIP collapses R@1 from 66.30\% to 10.35\%, and Table~\ref{tab:encoder_generality}
confirms that RS-specialized text encoders, despite exposure to satellite imagery,
degrade open-vocabulary performance relative to original CLIP, and the magnitude
of degradation scales with fine-tuning data volume.
GeoRSCLIP, fine-tuned on $\sim$5M optical RS pairs, outperforms RemoteCLIP
($\sim$830K pairs), yet both trail original CLIP (400M pairs): a larger fine-tuning
corpus partially cushions manifold compression, but even 5M RS pairs cannot
substitute for the semantic breadth distilled from 400M diverse examples.
This gradient carries a direct implication for SAR: the largest available SAR
caption datasets ($\sim$430K pairs) are an order of magnitude smaller than
RemoteCLIP's training corpus and nearly two orders of magnitude below GeoRSCLIP's,
placing SAR fine-tuning squarely in the most damaging region of this curve.
The fact that degradation persists even at 5M pairs, and intensifies at 830K,
empirically supports our central claim that SAR caption data is structurally
insufficient for fine-tuning, suggesting the frozen approach may be not merely
preferable but necessary in this regime.
On the SAR encoder side, unfreezing CROMA degrades mAP from 36.56\% to 15.27\%,
and the Procrustes analysis (Table~\ref{tab:procrustes}) quantifies this as a
structural geometric distortion, not merely a distributional shift.
Row B ($\Delta\!=\!0.351$) shows that fine-tuning CROMA on SAR captions distorts
its own learned manifold more than our bridge does (row A, $\Delta\!=\!0.210$),
a striking result: the bridge is \emph{more geometrically faithful to the original
SAR physics than supervised fine-tuning of CROMA itself}.
 
\noindent\textbf{Why cross-modal grounding matters.}
The backbone ablation (Table~\ref{tab:backbone_generality}) clarifies what
makes CROMA the right visual expert.
FG-MAE, pre-trained at the same scale on the same data, does not reach CROMA's
performance; we attribute this gap primarily to the absence of an optical anchor
rather than differences in capacity, since the two backbones share data, scale, and modality.
Without cross-modal contrastive alignment, the SAR encoder learns features that
are internally consistent but not structured to interface with the optical-grounded
CLIP text manifold.
This is consistent with the Platonic Representation
Hypothesis~\cite{pmlr-v235-huh24a}: it may be specifically the shared cross-modal
pre-training objective (CLIP: text$\leftrightarrow$optical; CROMA:
SAR$\leftrightarrow$optical) that induces compatible internal geometries,
rather than scale or architecture alone.
 
\noindent\textbf{Task scope and baseline comparisons.}
Our evaluation protocol is open-vocabulary zero-shot classification:
mapping imagery into an open semantic space of arbitrary text embeddings.
Unimodal SAR foundation models such as SARATR-X~\cite{10856784} are closed-set
feature extractors optimized for fixed label sets and cannot be directly evaluated
under this protocol without adding a supervised probe that changes the task entirely.
Our comparison set therefore consists of the current SOTA models that produce
SAR-to-text-space outputs: SAR-RS-CLIP and SARCLIP.
 
\noindent\textbf{Label efficiency.}
Our model outperforms SARCLIP ($\sim\!430$K pairs) and SAR-RS-CLIP ($\sim\!130$K
pairs) using only $\sim\!50$K pairs.
This $8\times$ reduction is consistent with the frozen design.
Because the encoders provide pre-built representations encoding SAR physics and
language semantics respectively, the bridge only needs to learn a geometric
rotation and scaling between two compatible manifolds, not reconstruct a full
cross-modal mapping from limited, noisy supervision.
The community's prevailing approach, scaling up synthetic SAR caption datasets, may
not be addressing the primary bottleneck.
The binding constraint appears to lie less in the volume of caption data than in
the quality and physical fidelity of the encoder manifolds.
 
\noindent\textbf{Implications and limitations.}
The principle demonstrated here is not specific to SAR.
Other domains where signal formation physics is incompatible with optical training
and where large-scale aligned caption data is unavailable are candidates
for this approach, including synthetic aperture lidar, acoustic imagery,
or hyperspectral modalities.
The key requirement is the availability of a domain-specific encoder pre-trained
with a cross-modal objective that grounds its representations in a modality
that shares semantic topology with the target language model.
Limitations include the upper bound imposed by the frozen experts: the bridge
cannot learn semantic concepts absent from the CLIP text manifold or CROMA's
visual feature space, limiting performance on rare or highly specialized
land-cover terminology.
The global average pooling representation also discards the spatial features
required for dense prediction tasks such as open-vocabulary segmentation.
Extending the Frozen Expert framework to pixel-level tasks via dense alignment
strategies, and investigating more recent SAR encoders as drop-in backbones,
are natural directions for future work.
 
\section{Conclusion}
 
We demonstrate that constrained alignment of frozen expert representations
outperforms full fine-tuning of baselines across a physical modality gap, mitigating
two distinct failure modes simultaneously: the corruption of the optical-pretrained
semantic manifold and the physical forgetting induced by gradient updates on
limited SAR caption data.
Our Frozen Expert framework bridges a frozen CROMA SAR encoder and a frozen CLIP
text encoder via a constrained, structure-preserving bridge with full mathematical
formulation (Eqs.~\ref{eq:bridge}--\ref{eq:infonce}).
We formally define physical forgetting via Procrustes disparity
(Eq.~\ref{eq:procrustes}), validate structure preservation empirically
(Table~\ref{tab:procrustes}), and demonstrate generality across multiple SAR
backbones and text encoders.
The result is SOTA zero-shot SAR scene classification across three benchmarks
at $8\times$ less training data than leading fine-tuned baselines.
More broadly, we expect the frozen-alignment principle to generalize to other
high-disparity cross-modal settings where specialized self-supervised pre-training
creates physically meaningful representations that would be erased by
fine-tuning on small, noisy paired datasets.
 
{\small
\bibliographystyle{ieeenat_fullname}
\bibliography{main}
}
 
\end{document}